\documentclass{article}

\usepackage{microtype}
\usepackage{graphicx}
\usepackage{subfigure}
\usepackage{booktabs} 

\usepackage{hyperref}

\newcommand{\amn}[1]{\prod_{i=1}^{#1}a_i}

\usepackage[accepted]{icml2024}

\usepackage{amsmath}
\usepackage{amssymb}
\usepackage{mathtools}
\usepackage{amsthm}

\usepackage[capitalize,noabbrev]{cleveref}

\theoremstyle{plain}
\newtheorem{theorem}{Theorem}[section]
\newtheorem{proposition}[theorem]{Proposition}

\theoremstyle{definition}
\newtheorem{definition}[theorem]{Definition}

\theoremstyle{remark}

\usepackage[textsize=tiny]{todonotes}

\icmltitlerunning{Probability-Generating Function Kernels for Spherical Data}

\begin{document}

\twocolumn[
\icmltitle{Probability-Generating Function Kernels for Spherical Data}




\begin{icmlauthorlist}
\icmlauthor{Theodore Papamarkou}{uom}
\icmlauthor{Alexey Lindo}{uog}
\end{icmlauthorlist}

\icmlaffiliation{uom}{Department of Mathematics, The University of Manchester, Manchester, United Kingdom}
\icmlaffiliation{uog}{School of Mathematics and Statistics, University of Glasgow, Glasgow, United Kingdom}

\icmlcorrespondingauthor{Theodore Papamarkou}{theo.papamarkou@manchester.ac.uk}
\icmlcorrespondingauthor{Alexey Lindo}{alexey.lindo@glasgow.ac.uk}

\icmlkeywords{Compositional kernels,
Gaussian processes,
deep kernel learning,
kernels,
neural network Gaussian processes,
probability generating functions,
spherical data}

\vskip 0.3in
]



\printAffiliationsAndNotice{}  

\begin{abstract}
Probability-generating function (PGF) kernels are introduced,
which constitute a class of kernels supported on the unit hypersphere,
for the purposes of spherical data analysis.
PGF kernels generalize RBF kernels in the context of spherical data.
The properties of PGF kernels are studied.
A semi-parametric learning algorithm is introduced
to enable the use of PGF kernels with spherical data.
\end{abstract}

\section{Introduction}


This paper contributes to spherical data analysis
by developing a new class of kernels, called
probability-generating function (PGF) kernels.
By construction, PGF kernels are supported on the unit hypersphere,
making them a natural choice for spherical data.


PGF kernels are dot-product kernels.
Furthermore, PGF kernels generalize
radial basis function (RBF) kernels,
and therefore the former are expected to have better
predictive performance on spherical data than the latter.


\textbf{Contribution.}
A summary of contributions follows.
PGF kernels are introduced as compositions of PGFs.
This way, another connection
between machine learning and probability is drawn.
It is shown that PGF kernels generalize RBF kernels.
A semi-parametric learning algorithm is introduced to
fit PGF kernels to spherical data.
PGF kernels are equipped with a notion of depth and width,
both of which are explored via an ablation study.
Three examples are presented to demonstrate the applicability of PGF kernels,
namely two Gaussian process (GP) regression examples
on circular and spherical data,
and a deep kernel learning (DKL) classification example
on higher-dimensional spherical data.



\textbf{Comparison with other works.}
\citet{daniely2016} have studied theoretical aspects of
compositional kernels based on generating functions (GFs),
with an emphasis on the duality between GF kernels
and (neural network) activation functions.
In contrast, this paper focuses on
PGF kernels (rather than GF kernels),
studying their properties and relation to other kernels,
and introducing a learning algorithm that enables
the use of PGF kernels in practice.
~\citet{liang2021} have focused on a subclass of PGF kernels,
based on compositions of a single PGF,
and have introduced a learning algorithm for this subclass
by drawing a link between kernels and activation functions.
In contrast, this paper defines PGF kernels in their generality,
allowing compositions of different PGFs,
and puts forward a learning algorithm that allows
the use of PGF kernels as a standalone entity,
without necessitating a link with activation functions.


\textbf{Paper structure.}
Section~\ref{sec:related_work}
highlights related work in the literature.
Section~\ref{sec:pgf_kernels}
introduces PGF kernels,
studies their properties and relations with other kernels,
and provides examples of kernels based on known PGFs.
Section~\ref{sec:learning_with_pgf_kernels}
introduces a learning algorithm to fit PGF kernels to spherical data,
and uses this algorithm for GP regression and DKL classification.
The paper concludes with a discussion,
providing directions for future work (section~\ref{sec:discussion}).

\section{Related Work}
\label{sec:related_work}


\textbf{Spherical data analysis.}
Spherical data analysis refers to the analysis of data supported on a hypersphere.
Numerous methods have been developed for spherical data analysis.
~\citet{lei2019} have introduced
a neural network architecture and a spherical convolutional kernel,
the latter being used as a component of the former.
The approach of~\citet{lei2019} is designed for point cloud data in $\mathbb{R}^3$,
which are typically processed using spherical regions.
~\citet{marques2022} have applied GP regression to approximate
incident radiance functions on the sphere in $\mathbb{R}^3$.
Several works have focused on the development of kernels on a hypersphere.
~\citet{guinness2016} have studied the differentiability properties of kernels on a hypersphere.
~\citet{pennington2015} have introduced a method to approximate
polynomial kernels for data on the unit sphere
using random spherical Fourier features.
~\citet{zhao2018} have proposed a power series expansion of the RBF kernel
for heat diffusion on a hypersphere.

\textbf{PFGs.}
A PGF is defined as
\begin{equation}
\label{eq:pgf_def}
g(s)
:= \sum_{i = 0}^{\infty} p_i s^i,
\end{equation}
where $p_{i} \in[0, 1],~i=0, 1, 2,\ldots$,
is a sequence of probabilities.
It is assumed that $\sum_{i \ge 0} p_i \le 1 $,
and therefore the power series in equation~\ref{eq:pgf_def}
converges absolutely for all $|s| \le 1$~\citep{grimmett2020}.
PGFs were introduced by De Moivre in 1730~\citep{seal1949,fischer2011}.
Nowadays, PGFs are widely used in branching processes,
theory of random walks,
renewal theory, and analytic combinatorics.





\textbf{Compositional kernels.}
Composition of kernels is a common operation in practice.
Python packages, such as \texttt{GPyTorch}~\citep{gardner2018},
provide functionality and examples of kernel composition.
The term `compositional kernel' is also linked
to the more topical class of kernels investigated by~\citet{daniely2016, liang2021},
establishing a duality between kernels and activation functions.
This paper has its methodological origin
in the topical notion of compositional kernels~\citep{daniely2016, liang2021},
but it shifts the attention away from their duality with activation functions.
In doing so, this paper is thus focused on the general operation of kernel composition,
introducing a methodological and computational framework to compose kernels using PGFs.


\textbf{GPs and DKL.}
Section~\ref{sec:learning_with_pgf_kernels} relies on GPs and DKL.
The log-marginal likelihoods used for GP regression and GP classification (as part of DKL)
are available in~\citet[equation 2.30]{rasmussen2005} and~\citet{milios2018}, respectively.
For DKL, the reader is referred to~\citet{wilson2016}.

\section{PGF Kernels}
\label{sec:pgf_kernels}

This section introduces PGF kernels,
their relations to other kernels,
and their properties.
It also provides examples of PGF kernels based on known PGFs.
Hereafter, $\rho (x, z) := \langle x, z \rangle$
denotes the correlation between $x\in\mathbb{S}^h$ and $z\in\mathbb{S}^h,$
where $\langle \cdot, \cdot \rangle$ is
the dot product and
$\mathbb{S}^h:=\{x\in\mathbb{R}^{h+1}: \|x\| = 1\}$
is the unit $h$-sphere with $h\ge 1$.

\subsection{Definition}

Definition~\ref{def:comp_kernel_in_varying_env} introduces PGF kernels.
PGF kernels apply compositions of PGFs on the correlation of their inputs.

\begin{definition}
\label{def:comp_kernel_in_varying_env}
Let $g_1,\ldots,g_n$ be PGFs and
$\rho (x, z)$ the correlation between $x\in\mathbb{S}^h$ and $z\in\mathbb{S}^h$.
The $n$-depth PGF kernel $\mathcal{K}\left[g_{1}, \ldots, g_{n}\right]$
is defined as
\begin{equation}
\label{eq:kn:long:ve}
\mathcal{K}\left[g_{1}, \ldots, g_{n}\right](x, z) :=
g_n\circ\dots\circ g_1
(\rho(x, z)).
\end{equation}
\end{definition}

Setting $g=g_1=\dots=g_n$ in equation~\ref{eq:kn:long:ve}
yields
the special case
\begin{equation}
\label{eq:kn:long}
\mathcal{K}\left[(g)_{n}\right](x, z):=
\mathcal{K}[\,\underbrace{g, \ldots, g}_\text{$n$ times}\,](x, z) .
\end{equation}
Equation~\ref{eq:kn:long}
yields the
$n$-fold composition of $g$,
which is also known as the $n$-fold iteration of $g$
and is commonly denoted by $g_{(n)}$.
$\mathcal{K}[(g)_{n}]$ represents
the PGF kernel defined by
iterations of a single PGF,
previously referred to as compositional kernel in ~\citet{liang2021}.
The composition $g_n\circ\dots\circ g_1$
of different PGFs $g_1,\ldots,g_n$
is known in the literature on branching processes
as an iterated function system in varying or random environment
~\cite{kozlov1976, kersting2017, alsmeyer2021},
but it has not been used in machine learning problems.
Definition~\ref{def:comp_kernel_in_varying_env}
introduces PGF kernels
as compositions of different PGFs $g_1,\ldots,g_n .$

Definition~\ref{def:pgf_kernel_layer_and_width}
introduces the notions of
layer and layer width for PGF kernels.
This way, PGF kernels have
layers, depth and width, similarly to neural networks.

\begin{definition}
\label{def:pgf_kernel_layer_and_width}
Let $\mathcal{K}\left[g_{1}, \ldots, g_{n}\right]$ be a PGF kernel.
The $i$-th PGF $g_i$ is called the $i$-th compositional layer or, shortly,
the $i$-th layer of the kernel.
The number $l_i$ of non-zero terms in the series expansion of PGF $g_i$
(see equation~\ref{eq:pgf_def})
is called the width of the $i$-th layer.
\end{definition}

An $n$-depth PGF kernel $\mathcal{K}\left[g_{1}, \ldots, g_{n}\right]$
is parameterized by the coefficients of its PGFs $g_1, \ldots, g_n$.
Note that each layer width $l_i$ can be finite or infinite.
If all layer widths $(l_1, \ldots, l_n)$ are finite,
then the kernel contains $\sum_{i=1}^{n} l_i$ parameters.
If at least one of the layer widths is infinite,
then the kernel contains an infinite number of parameters.




\subsection{Relations to Common Kernels}


PGF kernels are dot-product kernels.
A dot-product kernel $g(\rho(x, z))$ entails an arbitrary function $g$~\cite{smola2001},
while a PGF kernel employs a composition $g_n\circ\dots\circ g_1$ of PGFs as $g$.
In fact, a composition $g_n\circ\dots\circ g_1$ of PGFs is a PGF.
So, a PGF kernel is a dot-product kernel for which $g$ is a PGF.
This remark clarifies that a PGF kernel can alternatively be defined as
$g(\rho(x, z))$, where $g$ is a PGF;
instead, definition~\ref{def:comp_kernel_in_varying_env} is chosen
to emphasize the role of compositional depth in machine learning problems.

RBF kernels are PGF kernels.
In other words,
each RBF kernel can be expressed as $g(\rho(x, z))$,
where $g$ is a PGF.

\begin{proposition}
\label{prop:rbf_as_pgf}
RBF kernels are PGF kernels.
\end{proposition}

\begin{proof}
Consider the RBF kernel
\begin{equation}
\label{eq:rbf}
K(x, z) =
\exp{\left(-\frac{1}{2\sigma^2} \lVert x-z\rVert^2\right)},
\end{equation}
where $\sigma$ is the scalelength.
Recall the polarization identity:
\begin{equation}
\label{eq:polarization_identity}
\lVert x-z\rVert^2 =
\lVert x \rVert^2 + \lVert y \rVert^2 -2\langle x, z\rangle .
\end{equation}
Since $\lVert x\rVert = \lVert z\rVert=1$,
it follows from equation~\ref{eq:polarization_identity} that
\begin{equation}
\label{eq:simplified_norm}
\lVert x-z\rVert^2 =
2 (1 - \langle x, z\rangle) .
\end{equation}
Combining equations
~\ref{eq:rbf} and~\ref{eq:simplified_norm} yields
\begin{equation}
\label{eq:rbf_rho}
K(x, z) =
\exp{\left(\frac{1}{\sigma^2} (\rho (x, z) - 1)\right)} .
\end{equation}
By expanding the exponential function
as a power series around the correlation,
equation~\ref{eq:rbf_rho} becomes
\begin{align}
\label{eq:rbf_a_expr}
K(x, z)
&=
\sum_{i=0}^{\infty} p_i \rho^i(x, z),\\
\label{eq:rbf_a_coefs}
p_i
&=
\exp{\left(-\frac{1}{\sigma^2}\right)}
\frac{1}{i! \sigma^{2i}} .
\end{align}
Equation~\ref{eq:rbf_a_expr} is a PGF kernel
if and only if for any $j\in\mathbb{N}\cup\{0\}$
it holds that
$0 \le p_i \le 1$ and $\sum_{i=0}^{\infty} p_i = 1$.
It is obvious from equation~\ref{eq:rbf_a_coefs} that $p_i > 0$.
Moreover, combining equations
~\ref{eq:rbf_rho} and~\ref{eq:rbf_a_expr} gives
\begin{equation*}
\sum_{i=0}^{\infty} p_i \rho^i(x, z) =
\exp{\left(\frac{1}{\sigma^2} (\rho (x, z) - 1)\right)},
\end{equation*}
and setting $\rho (x, z) = 1$ leads to
$\sum_{i=0}^{\infty} p_i = 1$,
whence it follows that $p_i \le 1$.
\end{proof}

Polynomial kernels with positive coefficients
summing up to one
are PGF kernels.
In reverse,
PGF kernels with a finite number of positive coefficients
are polynomial kernels.
It follows directly from these observations
that neither polynomial kernels nor PGF kernels
are subsumed by each other.

\subsection{Properties}



Proposition~\ref{prop:pure_theta_k_eigensys}
establishes the fact that PGF kernels are positive-definite
and expresses their eigensystem in closed form.
Proposition~\ref{prop:pure_theta_k_eigensys}
is a special case of
the corresponding proposition of
~\citet{smola2001}
for the eigensystem of dot-product kernels.

\begin{proposition}
\label{prop:pure_theta_k_eigensys}
Let
$\mathcal{K}\left[g_{1}, \ldots, g_{n}\right]
:\mathbb{S}^h\times\mathbb{S}^h\rightarrow \left[-1, 1\right]$
be a PGF kernel.
\begin{enumerate}
\item $\mathcal{K}\left[g_{1}, \ldots, g_{n}\right]$
is positive-definite
on $\mathbb{S}^h\times\mathbb{S}^h $.
\item The eigenfunctions of
$\mathcal{K}\left[g_{1}, \ldots, g_{n}\right]$ are
the spherical harmonic functions $Y_{ij}$ of
degree
$i\in\{0\}\cup\mathbb{N}$
and
of order $j=1,\ldots,r(h, i),$
with associated eigenvalues
\begin{equation*}
\label{eq:lambda}
\lambda_{ij} =
\frac{2p_i\pi^{h/2}}{\Gamma(h/2)r(h, i)}
\end{equation*}
of multiplicity
\begin{equation*}
\label{eq:r}
r(h, i) =
\frac{2i + h - 2}{i}
\begin{pmatrix}
i + h - 3\\
i - 1
\end{pmatrix}.
\end{equation*}
$p_i$
are the probabilities of the
PGF $g_n\circ\dots\circ g_1$
and $\Gamma$ is the Gamma function.
\end{enumerate}
\end{proposition}

\begin{proof}
Since $\mathcal{K}$ is a PGF kernel,
it is a dot-product kernel.
This allows to invoke
~\citet{smola2001},
which completes the proof.
\end{proof}

As stated in proposition~\ref{prop:stationarity},
PGF kernels are rotationally stationary,
which means that they are stationary
with respect to rotations about the origin.
In the present context, rotations are bijections
that preserve the origin and spherical distance.
PGF kernels are rotationally stationary
due to the fact that
the correlation $\rho(x, z)$ between two points $x$ and $z$
corresponds to the cosine of the angle between $x$ and $z$,
as elaborated in the proof of proposition~\ref{prop:stationarity}.

\begin{proposition}
\label{prop:stationarity}
PGF kernels are rotationally stationary.
\end{proposition}

\begin{proof}
Let $\mathcal{K}$ be a PGF kernel.
The dot product 
between the inputs $x$ and $z$ of $\mathcal{K}$
is given by 
$\langle x, z\rangle =
\lVert x \rVert \lVert z \rVert \cos{\theta}$,
where $\theta$ denotes the angle between $x$ and $z$.
Since $\lVert x\rVert = \lVert z\rVert=1$,
it follows that
$\rho (x, z) = \cos{\theta}$.
Let $x^{\prime}$ and $z^{\prime}$
be rotations of $x$ and $z$.
Due to the applied rotation,
the angle between
$x^{\prime}$ and $z^{\prime}$ is $\theta$.
Thus,
$\rho (x^{\prime}, z^{\prime}) = \rho (x, z) $,
and therefore
$\mathcal{K}(x, z) = \mathcal{K}(x^{\prime}, z^{\prime})$
according to equation~\ref{eq:kn:long:ve}.
\end{proof}

Proposition~\ref{prop:stationarity}
generalizes to dot-product kernels.
More generally,
rotational stationarity is a common property
for other kernel families operating on spherical data.

PGF kernels based on PGFs
that map perfectly correlated spherical data to one
are correlation kernels (see proposition~\ref{prop:corr}).
In other words, such kernels map
any two points into $\left[-1, 1\right]$,
and any two perfectly correlated points to one.

\begin{proposition}
\label{prop:corr}
Let $g_1,\ldots,g_n$ be PGFs.
If for each $i=1,\ldots,n$, the PGF $g_i$ satisfies $g_i(1) = 1$,
then the PGF kernel $\mathcal{K}\left[g_{1}, \ldots, g_{n}\right]$
is a correlation kernel.
\end{proposition}

\begin{proof}
Let $x$ and $z$ be two points on the unit sphere $\mathbb{S}^h$.
For any PGF $g_i,~i=1,\ldots,n$, it holds that
$|g_i(s)| \le 1,~s\in\left[-1,1\right]$.
Furthermore, the correlation function satisfies $|\rho(x, z)| \le 1$.
Thus,
$|\mathcal{K}\left[g_{1}, \ldots, g_{n}\right](x, z)| \le 1$.
It is assumed that $g_i (1) = 1$ for all $i=1,\ldots,n$.
Since $\rho(x, x) = 1$,
it follows that
$\mathcal{K}\left[g_{1}, \ldots, g_{n}\right](x, x) = 1$.
\end{proof}




\subsection{Kernels Based on Known PGFs}
\label{subsec:kernels_with_known_pgfs}

A PGF can be expressed
as an infinite series according to equation~\ref{eq:pgf_def}
or as a closed-form function.
Consequently, the resulting PGF kernel is expressed
as an infinite series
or as a closed-form function.
In this subsection, common PGFs are used to
provide examples of PGF kernels in closed form.
Such examples are provided to spark interest for future research,
as PGF kernels arising from common PGFs have not been studied.

Two possible ways to classify PFGs and the resulting PGF kernels
are based on the existence of explicit PGF iterations and
the number of non-zero terms in the series expansion
of equation~\ref{eq:pgf_def}.
First, two classes of PGFs are induced,
depending on whether the PGFs have explicit iterations.
The resulting PGF kernels are said to satisfy the closure property
if the corresponding PGFs have explicit iterations.
Second, PGFs without explicit iterations
can have a finite or infinite series expansion,
whereas PGFs with explicit iterations
have an infinite series expansion.
Thus, PGF kernels satisfying the closure property
are not polynomial kernels.

Two examples of PGF kernels
that do not satisfy the closure property
are based on binomial and Poisson PGFs,
hereafter termed binomial and Poisson kernels.
The lack of closure property implies
that binomial and Poisson kernels
can not be expressed in closed form,
and are therefore approximated numerically.

In what follows,
PGF kernels that arise from compositions of
Harris, linear fractional and $\theta$ kernels are discussed.
As a prerequisite,
appendix~\ref{app:pgfs}
recaps on the definitions of
Harris, linear fractional and $\theta$ PGFs.
Appendix~\ref{app:pgf_kernels_with_closure}
states and proves the closure and other properties
of some of these kernels.

\begin{definition}
\label{def:harris_kernel}
Let $\rho (x, z)$ be the correlation between $x\in\mathbb{S}^h$ and $z\in\mathbb{S}^h$.
The Harris kernel with parameters $c$ and $r$ is defined as
\begin{equation}
\label{eq:harris_kernel}
\mathcal{H}_{c, r}(x, z)
:= \left(c  \rho (x, z)^{-r} - \left( c -1 \right) \right)^{-\frac{1}{r}},
\end{equation}
where
$r\in\mathbb{N}$
and
$c > 1$.
\end{definition}

\begin{definition}
\label{def:lf_kernel}
Let $\rho (x, z)$ be the correlation between $x\in\mathbb{S}^h$ and $z\in\mathbb{S}^h$.
The linear fractional kernel with parameters $a$ and $b$ is defined as
\begin{equation}
\label{eq:lf_kernel}
\mathcal{F}_{a, b}(x, z) :=
1 - \left(a
(1-\rho (x, z))^{-1}
+ b
\right)^{-1} ,
\end{equation}
where
$a>0,$ $b>0,$ and $a+b \ge 1.$
\end{definition}

Table~\ref{tab:thetaK}
in appendix~\ref{app:pgf_kernels_with_closure} shows
all nine possible cases of PGF kernels
that arise from $n$-fold iterations
of the nine cases of $\theta$ PGFs
(see definition~\ref{dfn:theta_pgfs} in appendix~\ref{app:pgfs}).
Note that the PGF kernels based on $\theta$ PGFs
generalize linear fractional kernels,
since the latter are derived
from the first three cases of Table~\ref{tab:thetaK}
by setting $\theta=1$.
A PGF kernel that arises from the $n$-fold composition of $\theta$ PGFs
is considered in propositions
~\ref{prop:theta_kernel_ex} and~\ref{prop:theta_kernel_ex_inf_depth}.
This kernel and its infinite-depth version
can be expressed in closed form according to propositions
~\ref{prop:theta_kernel_ex} and~\ref{prop:theta_kernel_ex_inf_depth}, respectively
(the proofs are available in appendix~\ref{app:pgf_kernels_with_closure}).
The PGF kernel appearing in propositions
~\ref{prop:theta_kernel_ex} and~\ref{prop:theta_kernel_ex_inf_depth}
is qualitatively different from Harris and linear fractional kernels,
because it arises from a composition,
rather than an iteration, of $\theta$ PGFs.

\begin{proposition}
\label{prop:theta_kernel_ex}
Consider the $n$-depth PGF kernel
$\mathcal{T}_{\theta, c}[g_1,\ldots,g_n]$
that arises from the $n$-fold composition of $\theta$ PGFs
$g_{i} (s) =1-((1-s)^{-\theta }+c_{i})^{-1/\theta }$,
where $c_{i} > 0$ for $i=1\,\ldots,n$.
This kernel experesses as
\begin{equation}
\label{eq:theta_kernel_ex}
\begin{aligned}
& \mathcal{T}_{\theta, c}[g_1,\ldots,g_n](x, z) = \\
& 1-((1-\rho(x, z))^{-\theta }+c)^{-1/\theta }, 
\end{aligned}
\end{equation}
where $c = \sum_{i = 1}^{n} c_{i}$.
\end{proposition}

\begin{proposition}
\label{prop:theta_kernel_ex_inf_depth}
The infinite-depth limit of the PGF kernel of proposition~\ref{prop:theta_kernel_ex}
is given by
\begin{equation}
\label{eq:theta_k_limit}
\begin{aligned}
& \lim_{n\rightarrow\infty}
\mathcal{T}_{\theta, c}[g_1,\ldots,g_n](x, z)  = \\
& 1 - \left((1 - \rho (x, z))^{-\theta} + c \right)^{-\frac{1}{\theta}},
\end{aligned}
\end{equation}
where
$c=\sum_{i = 1}^{\infty} c_{i}$.
\end{proposition}

All the PGF kernels mentioned in subsection~\ref{subsec:kernels_with_known_pgfs}
are correlation kernels.
However, arbitrary PGF kernels,
as used in learning tasks (section~\ref{sec:learning_with_pgf_kernels}),
are not necessarily correlation kernels.



\section{Learning with PGF Kernels}
\label{sec:learning_with_pgf_kernels}

This section introduces a learning algorithm
to fit PGF kernels to spherical data (subsection~\ref{subsec:algo}).
The proposed algorithm is put in use
to run GP regression (subsection~\ref{subsec:gp_reg})
and DKL classification (subsection~\ref{subsec:dkl}).




\subsection{Learning Algorithm}
\label{subsec:algo}



Fitting a model equipped with a PGF kernel to a dataset,
involves learning the kernel parameters,
which are the PGF coefficients.
In practice, the number of PGF parameters is reduced
by introducing a numerical approximation
to ensure numerical stability and computational tractability.
Consider an $n$-depth PGF kernel
$\mathcal{K}\left[g_{1}, \ldots, g_{n}\right]$
with layer widths $(l_1,\ldots,l_n)$.
Each PGF $g_i$ is numerically approximated by a series truncation.
More concretely, the first $m_i < \infty$ terms are maintained
in the series expansion of PGF $g_i$, where $m_i \le l_i$.
So, the numerical approximation of PGF kernel
$\mathcal{K}\left[g_{1}, \ldots, g_{n}\right]$
contains $m=\sum_{i=1}^{n}m_i$ parameters.
In this paper, $(m_1,\ldots,m_n)$
are referred to as (truncation) widths.

Typically, kernels are used in Bayesian non-parametrics.
While PGF kernels are suitable for non-parametric modeling,
they contain more parameters than other kernels.
From this perspective,
numerical approximations of PGF kernels introduce a semi-parametric setup.

To learn the $m$ PGF kernel parameters,
stochastic optimization is run on
a set of corresponding real-valued parameters.
More concretely,
at each iteration of the optimization algorithm,
a gradient descent step is taken in the space of real-valued parameters,
which are then transformed to approximate PGF coefficients
via the softmax function in order to evaluate the loss function.

\subsection{GP Regression}
\label{subsec:gp_reg}


\subsubsection{Circular von Mises Density}


This subsubsection provides examples based on the circular von Mises density
$f:\mathbb{S}^1\rightarrow [0,\infty)$ given by
\begin{equation*}
\label{eq:von_mises_pdf}
f(x \lvert \kappa, \mu) = 
\frac{\exp{(\kappa \cos{(x - \mu )}})}{2\pi I_0(\kappa)},
\end{equation*}
where
$\kappa$ and $\mu$ correspond to the shape and location parameters,
and $I_0 (\kappa)$ denotes
the modified Bessel function of the first kind of order zero.
In the experiments,
the von Mises density parameters are set to
$\kappa = 2$ and $\mu = (0, 0)^T$.
Two GP regression examples
related to the circular von Mises density are presented,
namely an ablation study of PGF kernel width and depth, and
a performance comparison between PGF and other kernels.

%

The ablation study aims to demonstrate the role of width and depth
in the predictive performance of PGF kernels.
To this end,
two sets of comparisons are run.
To assess the role of width,
three single-layer PGF kernels with increasing width
$m_1 = 2,~m_1=100$ and $m_1=200$ are compared.
To assess the role of depth,
three fixed-width PGF kernels with increasing depth
$n=1$, $n=2$ and $n=3$
are compared;
more specifically,
the widths are set to
$m_1=10$,
$(m_1, m_2)=(10, 10)$ and
$(m_1, m_2, m_3)=(10, 10, 10)$.

A training set is simulated
by drawing $250$ samples from the circular von Mises density
and adding Gaussian noise $\mathcal{N}(\mu=0, \sigma=0.5)$ to them.
For each PGF kernel,
a GP regression model is fitted to the training set
by running the Adam optimizer for the same number of iterations ($200$).
The standard log-marginal likelihood function for GP regression is used;
see~\citet[equation 2.30]{rasmussen2005}.
A test set of $250$ data points
is drawn from the circular von Mises density.
Using the estimated PGF kernel parameters
obtained from training,
predictions are made on the known ground truth,
that is, on the noiseless test set.

\begin{figure*}[!t]
\begin{center}
\centerline{\includegraphics[width=1.\linewidth]{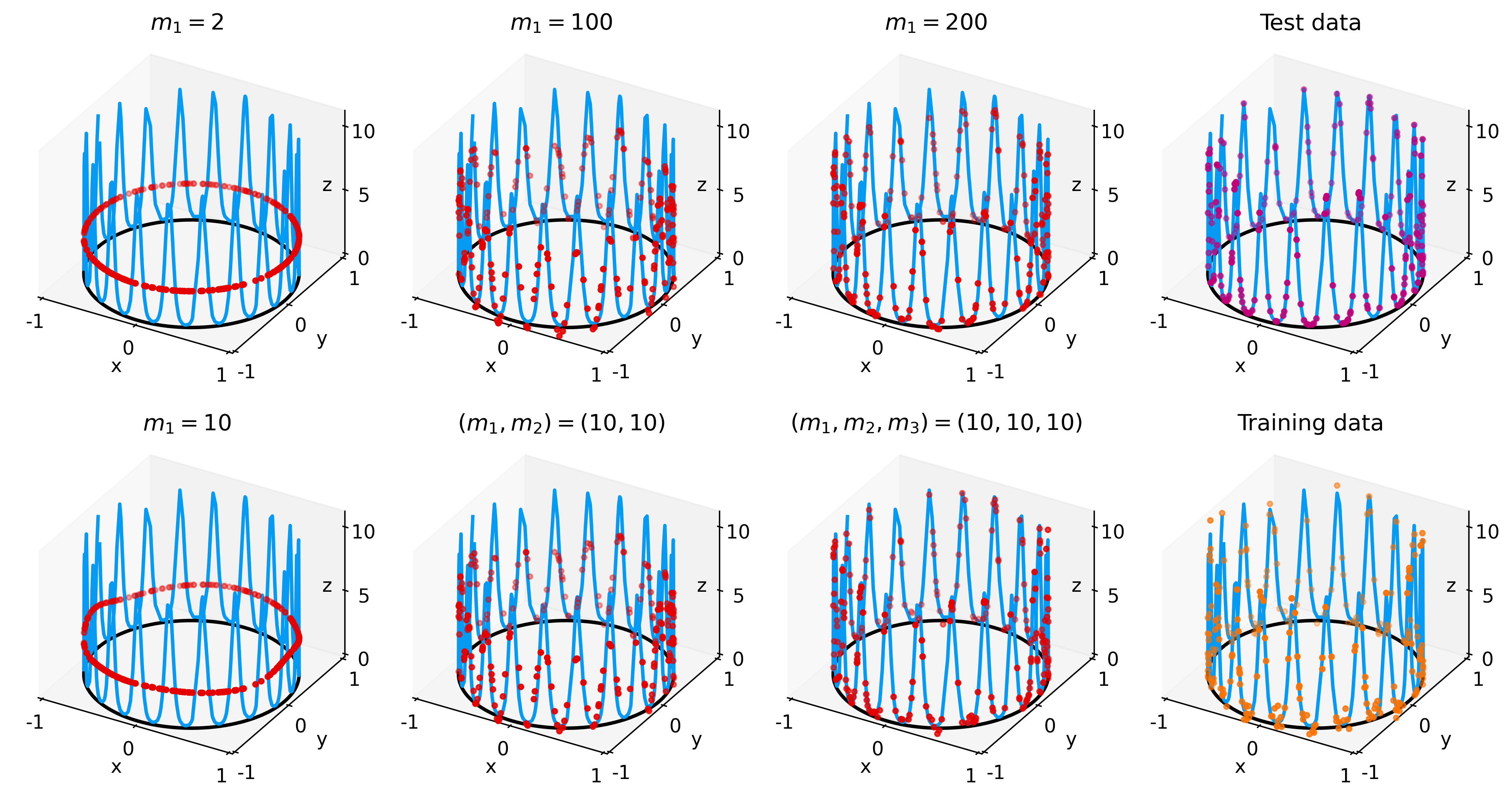}}
\caption{Visual demonstration of the effect of
PGF kernel width and depth on predictive performance
for GP regression fitted
to data drawn from the circular von Mises density.
The blue line represents the circular von Mises density
from which the data are simulated.
The orange, pink and red points represent
training data, test data and predictions, respectively.
Increasing the width or depth
brings the predictions (red points)
closer to the test data (pink points)
and therefore improves predictive performance.}
\label{fig:von_mises}
\end{center}
\vskip -0.3in
\end{figure*}

Figure~\ref{fig:von_mises}
visualizes the relative predictive performance
of PGF kernels of varying width or depth.
The blue line in each plot represents
the underlying circular von Mises density.
The orange points on the bottom right plot
and the pink points on the top right plot
correspond to the noisy training set
and the noiseless test set.
The red points in the other plots
of Figure~\ref{fig:von_mises}
depict predictions.
The first three plots in row one
correspond to single-layer PGF kernels
with widths $m_1 = 2,~m_1=100$ and $m_1=200$,
while the first three plots in row two
correspond to fixed-width PGF kernels
with depths $n=1$, $n=2$ and $n=3$.
As can be seen from the first and second rows of plots,
increasing the width or depth
produces predictions (red points)
in closer proximity to the test set
(pink points in the top right plot).
In other words, Figure~\ref{fig:von_mises}
shows that increasing the width or depth of a PGF kernel
results in higher predictive performance.

Table~\ref{tab:von_mises_ablation_study}
summarizes numerically the predictive performance
of the simulation visualized in Figure~\ref{fig:von_mises}.
More specifically,
two error metrics are reported,
namely the mean absolute error (MAE)
and the mean squared error (MSE).
As observed,
both MAE and MSE are
strictly decreasing functions
of width and depth.

\begin{table}[t]
\caption{Error metrics (MAE and MSE) demonstrate
the effect of PGF kernel width and depth
on predictive performance
for GP regression fitted to data
drawn from the circular von Mises density.
Increasing the width or depth
reduces these error metrics
and therefore improves predictive performance.}
\label{tab:von_mises_ablation_study}
\begin{center}
\begin{small}
\begin{sc}
\begin{tabular}{rrr}
\toprule
\multicolumn{3}{c}{Single layer, varying width} \\
\midrule
\multicolumn{1}{c}{Width} &
\multicolumn{1}{c}{MAE} &
\multicolumn{1}{c}{MSE} \\
\midrule
  2 & 2.8564 & 11.8405 \\
100 & 1.3331 &  2.7496 \\
200 & 0.5144 &  0.4973 \\
\midrule
\multicolumn{3}{c}{Fixed width, varying depth} \\
\midrule
\multicolumn{1}{c}{Depth} &
\multicolumn{1}{c}{MAE} &
\multicolumn{1}{c}{MSE} \\
\midrule  
  1 & 2.8576 & 11.9041 \\
  2 & 1.4035 &  3.1663 \\
  3 & 0.2530 &  0.1082 \\
\bottomrule
\end{tabular}
\end{sc}
\end{small}
\end{center}
\end{table}


Next, a predictive performance comparison
is made between the PGF kernel
and other well-known kernels.
In this example,
the PGF kernel with
depth $n=3$ and
widths $(m_1, m_2, m_3)=(30, 30, 30)$ is used.
Thus, the employed PGF kernel has $m=90$ parameters.
The comparison is carried out against
the RBF, Matern, periodic and spectral mixture kernel.

Ten training sets, each consisting of $500$ data points, are simulated
following the sampling procedure described in the ablation study.
For each kernel,
a GP regression model is fitted to each training set
by running the Adam optimizer for the same number of iterations ($1,000$).
The adopted log-marginal likelihood function
is mentioned in~\citet[equation 2.30]{rasmussen2005}.
Ten test sets, each consisting of $500$ data points,
are drawn from the circular von Mises density.
For each kernel, predictions are made
and error metrics are computed
on each test set.

Table~\ref{tab:von_misses_kernel_comp}
summarizes numerically the predictive performance
based on the kernels under comparison.
For each kernel,
the mean MAE and mean MSE
across the ten test sets
are tabulated.
For these means, standard errors are also reported.
In this GP regression experiment,
the PGF kernel attains lower MAE and MSE
than the RBF kernel.
This is an empirical confirmation
of the fact that PGF kernels generalize RBF kernels
(see proposition~\ref{prop:rbf_as_pgf}).
Besides, the PGF kernel yields
the smallest means and standard errors for MAE and MSE
in comparison to the other kernels used in the experiment.

\begin{table}[!t]
\caption{Mean MAEs and MSEs (accompanied by standard errors)
for GP regression
fitted to data, drawn from the circular von Mises density,
using PGF, RBF, Matern, periodic and spectral mixture kernels.
For each kernel, the error metric
means and standard errors are obtained from
ten runs (for details, see the main text).
Metrics reported in bold indicate best performance.}
\label{tab:von_misses_kernel_comp}
\begin{center}
\begin{small}
\begin{sc}
\begin{tabular}{lrr}
\toprule
\multicolumn{1}{c}{Kernel} &
\multicolumn{1}{c}{MAE} &
\multicolumn{1}{c}{MSE} \\
\midrule
PGF      & \textbf{0.1683}$\pm$0.0181 & \textbf{0.0501}$\pm$0.0122 \\
RBF      & 1.3144$\pm$1.3583 & 4.5370$\pm$5.7027 \\
Matern   & 0.3766$\pm$0.0360 & 0.2621$\pm$0.0669 \\
Periodic & 0.3031$\pm$0.0265 & 0.1748$\pm$0.0289 \\
Spectral & 0.2553$\pm$0.0328 & 0.1170$\pm$0.0360 \\
\bottomrule
\end{tabular}
\end{sc}
\end{small}
\end{center}
\vskip -0.16in
\end{table}

\subsubsection{Spherical Exponential-Cosine Function}


This subsubsection provides
an example based on spherical input data living on $\mathbb{S}^2$
and output data simulated from an exponential-cosine (exp-cos) function.
More concretely, points on the sphere are simulated in polar coordinates $(\theta, \phi)$,
where $\theta$ and $\phi$ denote the polar angle and azimuthal angle, respectively.
The polar coordinates are transformed to Cartesian coordinates,
and the latter constitute the input data.
The output data are generated from the
spherical exp-cos function
$f: [0, \pi] \times [-\pi, \pi]\rightarrow [0,\infty)$
given by
\begin{equation}
\label{eq:expcosf}
f(\theta, \phi) =
\exp{
\left(
u\sum_{i=1}^{4}
\cos{
\left(
v (a_i \theta + b_i \phi + c_i)
\right)
}
\right)
},
\end{equation}
where
$(a_1, a_2, a_3, a_4) = (0, 1, 1, 1)$,
$(b_1, b_2, b_3, b_4) = (1, 1, 0, 1)$,
$(c_1, c_2, c_3, c_4) = (0, \pi/2, \pi, 3\pi/2)$,
$u = 0.5$ and $v = 15$.

A predictive performance comparison
is made between the PGF kernel and
the RBF, Matern, periodic and spectral mixture kernel.
In this example,
the PGF kernel with
depth $n=3$ and
widths $(m_1, m_2, m_3)=(20, 20, 20)$ is used,
which has $m=60$ parameters.

Four training sets, each consisting of $5,000$ data points, are generated.
For each training set,
$5,000$ input data points are simulated by sampling uniformly at random
the polar angle $\theta$ and azimuthal angle $\phi$. 
The corresponding output data points are generated by
evaluating the exp-cos function of equation~\ref{eq:expcosf}
at the input data points and by adding Gaussian noise
$\mathcal{N}(\mu = 0, \sigma=0.2)$.
For each kernel,
a GP regression model is fitted to each training set
by running the Adam optimizer for the same number of iterations ($1,000$).
The adopted log-marginal likelihood function
is available in~\citet[equation 2.30]{rasmussen2005}.
Four noiseless test sets, each consisting of $5,000$ data points,
are generated similarly to the training sets,
the only difference being that noise is not added to the output.
The test sets are generated without noise
to ensure that predictive performance is evaluated
against the known ground truth given by equation~\ref{eq:expcosf}.
For each kernel, predictions are made
and error metrics are computed
on each test set.

\begin{figure}[!t]
\begin{center}
\centerline{\includegraphics[width=\columnwidth]{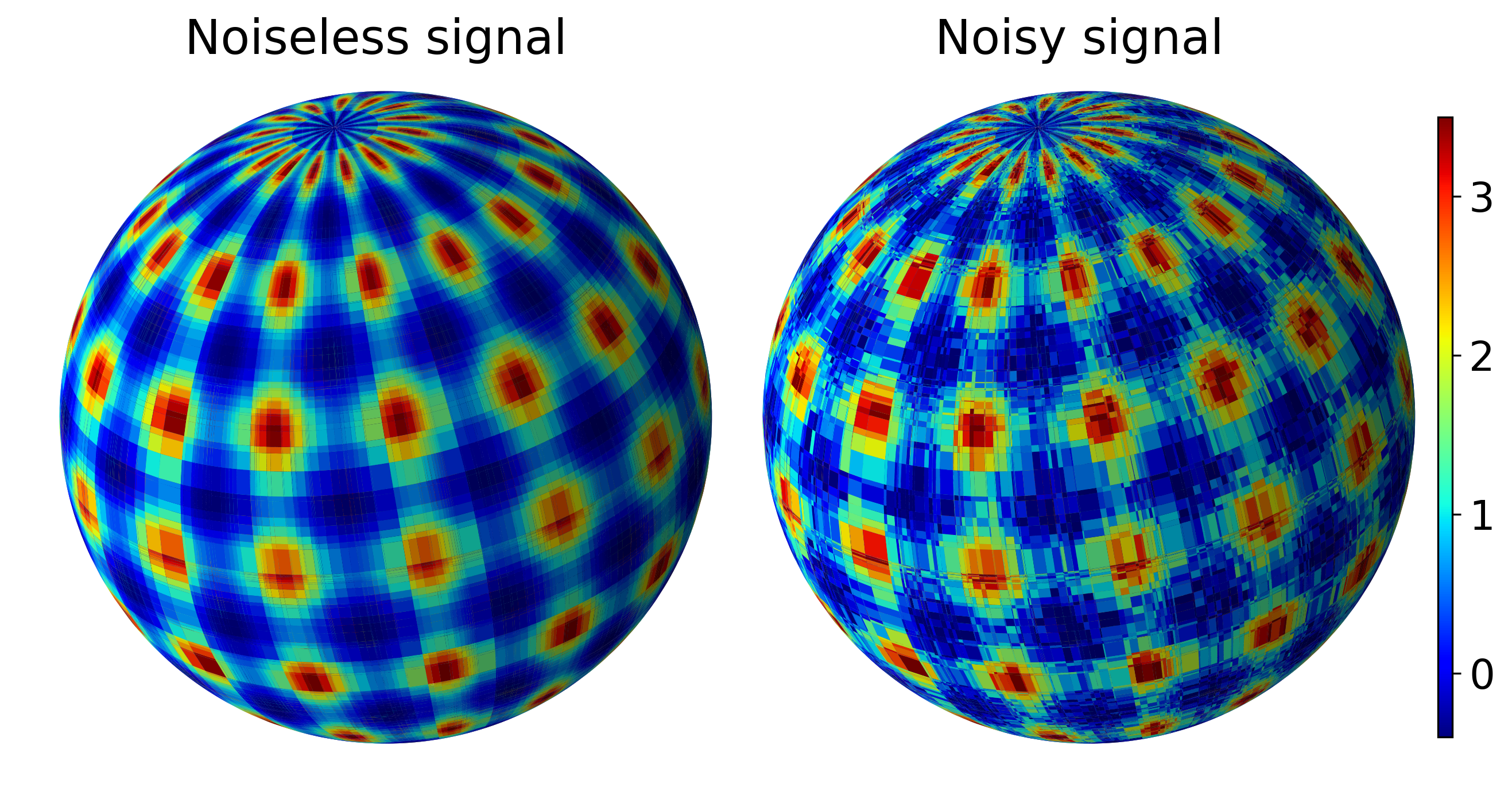}}
\caption{Noiseless (left) and
noisy version (right) of the spherical exp-cos function
given by equation~\ref{eq:expcosf}.
For the noisy version,
Gaussian noise $\mathcal{N}(\mu = 0, \sigma=0.2)$
is added to the exp-cos function.
Training sets are drawn from such noisy realizations,
whereas test sets are drawn from the noiseless version of the exp-cos function.}
\label{fig:discoball}
\end{center}
\vskip -0.6in
\end{figure}

Figure~\ref{fig:discoball}
demonstrates the training and test set generation.
The sphere on the left-hand side displays the exp-cos function
of equation~\ref{eq:expcosf},
from which test points are drawn.
The sphere on the right-hand side displays a noisy version of
the exp-cos function,
with noise drawn from
$\mathcal{N}(\mu = 0, \sigma=0.2)$;
training points are simulated from such realizations.

Table~\ref{tab:discoball_kernel_comp}
provides an empirical comparison of the kernels involved by
showing numerical summaries of predictive performance.
For each kernel,
the mean MAE, mean MSE and associated standard errors
across the four test sets
are tabulated.
In this GP regression experiment,
the PGF kernel outperforms the RBF kernel
in terms of MAE and MSE.
This observation confirms empirically
proposition~\ref{prop:rbf_as_pgf},
according to which PGF kernels form a superset of RBF kernels.
Moreover, the PGF kernel attains
the lowest MAE and second lowest MSE across all kernels.
The observed advantage of the PGF kernel
is supported by the different scaling of error metrics,
indicated by the reported standard errors.
Notably, only the PGF and Matern kernels
exhibit comparable predictive performance,
outperforming the RBF, periodic and spectral mixture kernel.

\begin{table}[!t]
\caption{Mean MAEs and MSEs (accompanied by standard errors)
for GP regression
fitted to data, drawn from the spherical exp-cos function
(equation~\ref{eq:expcosf}),
using PGF, RBF, Matern, periodic and spectral mixture kernels.
For each kernel, the error metric
means and standard errors are obtained from
four runs (for details, see the main text).
Metrics reported in bold indicate best performance.}
\label{tab:discoball_kernel_comp}
\begin{center}
\begin{small}
\begin{sc}
\begin{tabular}{lrr}
\toprule
\multicolumn{1}{c}{Kernel} &
\multicolumn{1}{c}{MAE} &
\multicolumn{1}{c}{MSE} \\
\midrule
PGF      & \textbf{0.1476}$\pm$0.0076 & 0.0443$\pm$0.0085 \\
RBF      & 0.3094$\pm$0.1598 & 0.1865$\pm$0.1444 \\
Matern   & 0.1504$\pm$0.0011 & \textbf{0.0442}$\pm$0.0033 \\
Periodic & 0.2500$\pm$0.0088 & 0.1149$\pm$0.0078 \\
Spectral & 0.1549$\pm$0.0106 & 0.0533$\pm$0.0075 \\
\bottomrule
\end{tabular}
\end{sc}
\end{small}
\end{center}
\vskip -0.4in
\end{table}

\subsection{Deep Kernel Learning}
\label{subsec:dkl}




As a last example, the supervised learning problem of
multiclass classification on the hypersphere is considered.
To this end, the hyperspherical Thomas process~\citep{thomas1949}
on $\mathbb{S}^{h}$ is used.
The hyperspherical Thomas process is conditioned on the number of clusters,
given the supervised learning setup.

Data are generated from this doubly stochastic process.
First, cluster centers are simulated on the hypersphere uniformly at random.
Second, for each cluster, data points are simulated
from a von Mises-Fisher distribution
whose mean direction is set to be the cluster center.
The number of data points per cluster is drawn from a Poisson distribution.
Thus, three sets of parameters are involved in the data generation process,
namely the intensity $\lambda$ of the underlying Poisson process,
and the mean directions $\mu$ and concentration parameters $\kappa$
of the von Mises-Fisher distributions.
Four datasets are simulated this way from
the hyperspherical Thomas process.
Each of the four datasets is split into two halves
to generate four corresponding training and test sets.
In this example, 
the multiclass classification problem involves four clusters,
and the parameters are set to
$h=17$,
$\lambda=850$ and $\kappa=(20, 20, 20, 20)$.
To give an indication of sample size,
the mean number of data points across the four datasets
(before being split into training and test sets)
is $8147$.


\begin{figure*}[!t]
\begin{center}
\centerline{\includegraphics[width=1.\linewidth]{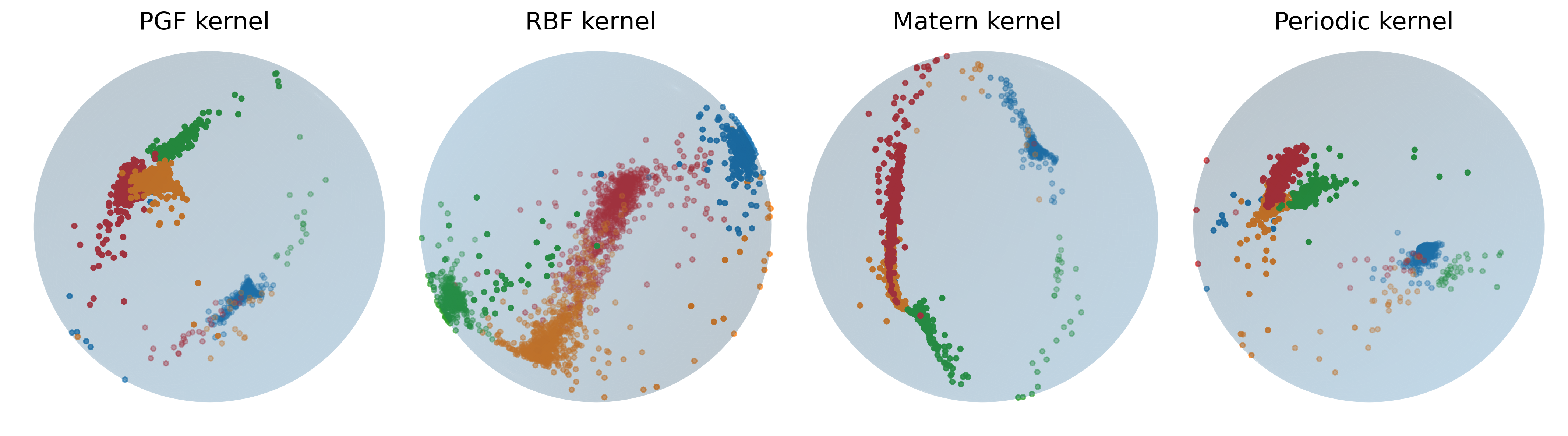}}
\caption{Embeddings in $\mathbb{R}^3$
generated by training a DKL classification model
on a dataset simulated from a hyperspherical Thomas process in $\mathbb{R}^{18}$.
Each of the four spheres displays the embeddings generated
by training the DKL model with a different kernel.
On each sphere, the four colors of the embeddings represent
the four classes of the classification problem.}
\label{fig:cox_proc}
\end{center}
\vskip -0.22in
\end{figure*}

A DKL classification model is fitted
to the data simulated from the hyperspherical Thomas process.
The adopted model is a modification
of the DKL framework introduced by~\citet{wilson2016}.
More specifically, the multilayer perceptron of~\citet{liang2021}
is used as the neural network component of the DKL model.
This choice of representation for the multilayer perceptron
is made to map the input data on the unit hypersphere
to embeddings on the unit sphere.
The embeddings must be on the unit sphere
to respect the input requirements for the PGF kernel.
The Dirichlet-based GP of~\citet{milios2018}
is employed as the GP component of the DKL model.
Thus, GP classification is performed
through the latter part of the DKL model,
using the embeddings as input to the GP.
This modified DKL model
has been chosen over standalone GP classification
for two reasons.
First, the DKL model reduces the computational cost of GP classification
by embedding the input into a lower-dimensional latent spherical space.
Second, these embeddings facilitate explainability,
as they constitute latent features in $\mathbb{R}^3$
that can be visualized.

A predictive performance comparison
is made between the PGF kernel and
the RBF, Matern and periodic kernel
in the context of this DKL example.
The PGF kernel with
depth $n=3$ and
widths $(m_1, m_2, m_3)=(20, 20, 20)$ is used.

For each kernel, two steps are taken.
First, a DKL model is fitted to each training set
by running the Adam optimizer for the same number of iterations ($500$).
Second, predictions are made
and error metrics are computed on the corresponding test set.

Figure~\ref{fig:cox_proc}
shows the embeddings generated
by fitting the DKL model to one of the training sets.
The embeddings produced by each kernel are visualized on the unit sphere.
On each displayed sphere, the four colors of the embeddings represent
the four classes of the classification problem.
The PGF kernel yields more distinct clusters of embeddings than the RBF kernel,
and more generally than the other kernels.

Table~\ref{tab:cox_proc_kernel_comp}
summarizes the predictive performance
of the four compared kernels.
For each kernel, the table shows
the mean predictive accuracy and its standard error
across the four test sets.
In this DKL classification experiment,
the PGF kernel outperforms the RBF kernel
in terms of predictive accuracy.
Moreover, the PGF kernel achieves the highest predictive accuracy
among the four kernels.

\begin{table}[!t]
\caption{Mean predictive accuracies and standard errors
for DKL cassification
fitted to data, drawn from the hyperspherical Thomas process,
using PGF, RBF, Matern and periodic kernels.
For each kernel, the mean predictive accuracy
and standard error are obtained from
four runs (for details, see the main text).
The bold font indicates best performance.}
\label{tab:cox_proc_kernel_comp}
\begin{center}
\begin{small}
\begin{sc}
\begin{tabular}{lr}
\toprule
\multicolumn{1}{c}{Kernel} &
\multicolumn{1}{c}{Accuracy (\%)} \\
\midrule
PGF      & \textbf{88.55}$\pm$4.8772 \\
RBF      & 88.30$\pm$5.1280 \\
Matern   & 88.10$\pm$5.0961 \\
Periodic & 88.27$\pm$4.9245 \\
\bottomrule
\end{tabular}
\end{sc}
\end{small}
\end{center}
\vskip -0.7in
\end{table}

So, the predictive performance of table~\ref{tab:cox_proc_kernel_comp}
and the clustering visualization of figure~\ref{fig:cox_proc}
demonstrate the superiority of the PGF kernel in this DKL example.
However, the standard errors of table~\ref{tab:cox_proc_kernel_comp}
indicate that the observed differences in performance between the four kernels
seem negligible.

\section{Discussion and Future Directions}
\label{sec:discussion}




This paper has introduced the family of PGF kernels.
Due to their parameterization,
PGF kernels replace kernel hyperparameter optimization
with a semi-parametric learning task.
Since PGF kernels generalize RBF kernels,
the former provide an alternative to the latter
in the case of spherical data.

Several directions for theoretical research arise.
PGF kernels satisfying the closure property may be used
to approximate arbitrary PGF kernels;
characterizing the quality of such an approximation
is an open problem.
This approximation is beneficial in two ways.
First, PGF kernels with or without the closure property
lead to non-parametric or semi-parametric learning tasks,
respectively.
For example, the fewer parameters of a PGF kernel
that satisfies the closure property
can be treated as hyperparameters in a non-parametric setting.
Thus, PGF kernels with the closure property,
if used as proxies to PGF kernels without the closure property,
can simplify the learning task.
Second, the closure property
substantially reduces the computational cost.

Another theoretical problem is to identify conditions
on the parameters of a PGF kernel
so that the kernel is optimal
for certain classes of data or learning tasks.
In other words, the question becomes how to select an optimal PGF
for a given dataset under the PGF kernel setup.

Spherical data analysis gives rise to future applications of PGF kernels.
For instance, simulations of partial differential equations on a sphere,
such as Boussinesq convection on a spherical shell or
viscous shallow-water motion on a sphere,
are computationally expensive.
GP emulators with PGF kernels can be used to reduce computational cost.
As another example, PGF kernels can find applications in earth data science,
including air temperature forecasting and ocean surface topography interpolation.






\section*{Acknowledgements}

This material is based upon work supported by
the Google Cloud Research Credits program
with the award GCP19980904.
The authors would like to thank
Devanshu Agrawal,
Serik Sagitov and
Umberto Noe
for useful discussions.



\begin{thebibliography}{21}
\providecommand{\natexlab}[1]{#1}
\providecommand{\url}[1]{\texttt{#1}}
\expandafter\ifx\csname urlstyle\endcsname\relax
  \providecommand{\doi}[1]{doi: #1}\else
  \providecommand{\doi}{doi: \begingroup \urlstyle{rm}\Url}\fi

\bibitem[Alsmeyer(2021)]{alsmeyer2021}
Alsmeyer, G.
\newblock Linear fractional {G}alton–{W}atson processes in random environment and perpetuities.
\newblock \emph{Stochastics and Quality Control}, 2021.

\bibitem[Daniely et~al.(2016)Daniely, Frostig, and Singer]{daniely2016}
Daniely, A., Frostig, R., and Singer, Y.
\newblock Toward deeper understanding of neural networks: the power of initialization and a dual view on expressivity.
\newblock In \emph{Advances in Neural Information Processing Systems}, volume~29. Curran Associates, Inc., 2016.

\bibitem[Fischer(2011)]{fischer2011}
Fischer, H.
\newblock \emph{A history of the central limit theorem: From classical to modern probability theory}.
\newblock Springer, 2011.

\bibitem[Gardner et~al.(2018)Gardner, Pleiss, Bindel, Weinberger, and Wilson]{gardner2018}
Gardner, J.~R., Pleiss, G., Bindel, D., Weinberger, K.~Q., and Wilson, A.~G.
\newblock G{P}y{T}orch: Blackbox matrix-matrix {G}aussian process inference with {GPU} acceleration.
\newblock In \emph{Advances in Neural Information Processing Systems}, 2018.

\bibitem[Grimmett \& Stirzaker(2020)Grimmett and Stirzaker]{grimmett2020}
Grimmett, G. and Stirzaker, D.
\newblock \emph{Probability and random processes}.
\newblock Oxford university press, fourth edition, 2020.

\bibitem[Guinness \& Fuentes(2016)Guinness and Fuentes]{guinness2016}
Guinness, J. and Fuentes, M.
\newblock Isotropic covariance functions on spheres: Some properties and modeling considerations.
\newblock \emph{Journal of Multivariate Analysis}, 143:\penalty0 143--152, 2016.

\bibitem[Harris(1963)]{harris1972}
Harris, T.~E.
\newblock \emph{The theory of branching processes}.
\newblock Springer-Verlag, 1963.

\bibitem[Kersting \& Vatutin(2017)Kersting and Vatutin]{kersting2017}
Kersting, G. and Vatutin, V.
\newblock \emph{Discrete time branching processes in random environment}.
\newblock John Wiley and Sons, 2017.

\bibitem[Kozlov(1977)]{kozlov1976}
Kozlov, M.~V.
\newblock On the asymptotic behavior of the probability of non-extinction for critical branching processes in a random environment.
\newblock \emph{Theory of probability and its applications}, 21\penalty0 (4):\penalty0 791--804, 1977.

\bibitem[Lei et~al.(2019)Lei, Akhtar, and Mian]{lei2019}
Lei, H., Akhtar, N., and Mian, A.
\newblock Octree guided {CNN} with spherical kernels for 3{D} point clouds.
\newblock In \emph{2019 IEEE/CVF Conference on Computer Vision and Pattern Recognition (CVPR)}, pp.\  9623--9632, 2019.

\bibitem[Liang \& Tran-Bach(2021)Liang and Tran-Bach]{liang2021}
Liang, T. and Tran-Bach, H.
\newblock Mehler’s formula, branching process, and compositional kernels of deep neural networks.
\newblock \emph{Journal of the American Statistical Association}, pp.\  1--14, 2021.

\bibitem[Marques et~al.(2022)Marques, Bouville, and Bouatouch]{marques2022}
Marques, R., Bouville, C., and Bouatouch, K.
\newblock Gaussian process for radiance functions on the $\mathbb{S}^2$ sphere.
\newblock \emph{Computer Graphics Forum}, 41\penalty0 (6):\penalty0 67--81, 2022.

\bibitem[Milios et~al.(2018)Milios, Camoriano, Michiardi, Rosasco, and Filippone]{milios2018}
Milios, D., Camoriano, R., Michiardi, P., Rosasco, L., and Filippone, M.
\newblock Dirichlet-based {G}aussian processes for large-scale calibrated classification.
\newblock In \emph{Proceedings of the 32nd International Conference on Neural Information Processing Systems}, NIPS'18, pp.\  6008--6018. Curran Associates Inc., 2018.

\bibitem[Pennington et~al.(2015)Pennington, Yu, and Kumar]{pennington2015}
Pennington, J., Yu, F. X.~X., and Kumar, S.
\newblock Spherical random features for polynomial kernels.
\newblock In \emph{Advances in Neural Information Processing Systems}, volume~28. Curran Associates, Inc., 2015.

\bibitem[Rasmussen \& Williams(2005)Rasmussen and Williams]{rasmussen2005}
Rasmussen, C.~E. and Williams, C.~K.
\newblock \emph{Gaussian processes for machine learning}.
\newblock MIT Press, 2005.

\bibitem[Sagitov \& Lindo(2016)Sagitov and Lindo]{sagitov2016}
Sagitov, S. and Lindo, A.
\newblock \emph{A special family of {G}alton-{W}atson processes with explosions}, pp.\  237--254.
\newblock Springer International Publishing, 2016.

\bibitem[Seal(1949)]{seal1949}
Seal, H.~L.
\newblock The historical development of the use of generating functions in probability theory.
\newblock \emph{Mitt Verein Schweiz Versich Math}, 49:\penalty0 209--228, 1949.

\bibitem[Smola et~al.(2001)Smola, \'{O}v\'{a}ri, and Williamson]{smola2001}
Smola, A., \'{O}v\'{a}ri, Z., and Williamson, R.~C.
\newblock Regularization with dot-product kernels.
\newblock In \emph{Advances in Neural Information Processing Systems}, volume~13. MIT Press, 2001.

\bibitem[Thomas(1949)]{thomas1949}
Thomas, M.
\newblock A generalization of {P}oisson's binomial limit for use in ecology.
\newblock \emph{Biometrika}, 36\penalty0 (1/2):\penalty0 18--25, 1949.

\bibitem[Wilson et~al.(2016)Wilson, Hu, Salakhutdinov, and Xing]{wilson2016}
Wilson, A.~G., Hu, Z., Salakhutdinov, R., and Xing, E.~P.
\newblock Deep kernel learning.
\newblock In \emph{Proceedings of the 19th International Conference on Artificial Intelligence and Statistics}, volume~51 of \emph{Proceedings of Machine Learning Research}, pp.\  370--378. PMLR, 2016.

\bibitem[Zhao \& Song(2018)Zhao and Song]{zhao2018}
Zhao, C. and Song, J.~S.
\newblock Exact heat kernel on a hypersphere and its applications in kernel {SVM}.
\newblock \emph{Frontiers in Applied Mathematics and Statistics}, 4, 2018.

\end{thebibliography}
\bibliographystyle{icml2024}

\newpage
\appendix
\onecolumn

\section{PGFs with Explicit Iterations}
\label{app:pgfs}

This appendix recaps on the definitions of
Harrris, linear fractional and $\theta$ PGFs.
Definition \ref{dfn:theta_pgfs} reproduces the definition of
$\theta$ PGFs, as introduced in~\citet{sagitov2016}.


\begin{definition}
\label{def:harris_pgf}
A Harris PGF with parameters $c$ and $r$ is defined as
\begin{equation}
\label{eq:harris_pgf}
g(s) = \left(c s^{-r} - (c - 1)\right)^{-\frac{1}{r}},~s\in [-1, 1],
\end{equation}
where
$r\in\mathbb{N}$
and
$c > 1$.
\end{definition}


\begin{definition}
A linear fractional PGF with parameters $a$ and $b$ is defined as
\begin{equation}
\label{eq:lf_pgf}
g(s) =
1 - \left(a
(1-s)^{-1}
+ b
\right)^{-1},~s\in [-1, 1],
\end{equation}
where
$a>0,$ $b>0,$ and $a+b \ge 1.$
\end{definition}


\begin{definition}
\label{dfn:theta_pgfs}
For $\theta \in (-1, 0) \cup (0, 1]$, consider
the PGF
\begin{equation}\label{eq:PGFMain}
g(s) =
r-(a(r-s)^{-\theta}+c)^{-1/\theta}, \theta\in (-1, 0) \cup (0, 1] .
\end{equation}
It is assumed that the parameters $\theta$, $a$, $c$ and $r$
satisfy one of the three options
\begin{equation*}
\begin{array}{llllll}
\theta\in(0,1], & a\ge1,  & c>0, & & r=1,\\
\theta \in (-1, 0) \cup (0, 1], & a\in(0,1),&  c=(1-a) (1-q)^{-\theta}, &q\in[0,1), & r=1,\\
\theta \in (-1, 0) \cup (0, 1], & a\in(0,1),&  c=(1-a) (r-q)^{-\theta}, &q\in[0,1], & r>1.
\end{array}
\end{equation*}

For $\theta=0$, consider the PGF
\begin{equation}\label{eq:PGFZero}
g(s) =
r-(r-q)^{1-a}(r-s)^{a}, \theta = 0.
\end{equation}
It is assumed that parameters $q$ and $r$ satisfy one of the two options
\begin{equation*}
\begin{array}{ll}
q \in [0, 1), & r = 1, \\
q \in [0, 1], & r > 1.
\end{array}
\end{equation*}

For $\theta=-1$, consider the PGF
\begin{equation}\label{eq:PGFMinus}
g(s) =
as+(1-a)q, \theta = - 1 .
\end{equation}
It is assumed that $a \in (0, 1)$ and $q \in [0, 1]$.

For each of the three cases
$\theta \in (-1, 0) \cup (0, 1]$,
$\theta=0$ and $\theta=-1$,
the corresponding PGF given by
equation~\ref{eq:PGFMain},~\ref{eq:PGFZero} and~\ref{eq:PGFMinus}
is called a $\theta$ PGF.
\end{definition}

For any fixed choice of $\theta$, $a$, $c$ and $r$,
an $n$-fold iteration of the corresponding $\theta$ PGF
yields a $\theta$ PGF.
Thus, an $n$-fold iteration of a $\theta$ PGF is expressed in closed form.
The families of $\theta$ PGFs and of Harris PGFs
~\citep[p. 10]{harris1972}
are the only two known PGF families
with explicit $n$-fold iterations.

\section{PGF Kernels Satisfying the Closure Property}
\label{app:pgf_kernels_with_closure}

This appendix establishes some properties of
PGF kernels that arise from compositions
of Harris, linear fractional and $\theta$ kernels.
Proposition~\ref{prop_harris_kernel_symm} states
a symmetry property of the Harris kernel.
Propositions~\ref{prop:harris_comp},~\ref{prop:lf_comp} and~\ref{prop:theta_kernels}
state the corresponding closure property of
the Harris kernel, linear fractional kernel
and PGF kernels arising
from $n$-fold iterations of $\theta$ PGFs.
The appendix concludes with the proofs
of propositions~\ref{prop:theta_kernel_ex} and~\ref{prop:theta_kernel_ex_inf_depth}.


\begin{proposition}
\label{prop_harris_kernel_symm}
For any
even $r$ and any
$(x,z)\in \mathbb{S}^h \times \mathbb{S}^h,$
the Harris kernel safisfies
\begin{equation}
\label{eq:harris_kernel_symm}
\mathcal{H}_{c,r}(x, z) =
\mathcal{H}_{c,r}(-x, z) =
\mathcal{H}_{c,r}(x, -z) .
\end{equation}
\end{proposition}

\begin{proof}
Equation~\ref{eq:harris_kernel_symm}
follows from equation~\ref{eq:harris_kernel} for even $r .$
\end{proof}

\begin{proposition}
\label{prop:harris_comp}
For any
$n\in\mathbb{N},$
let
$\mathcal{K}\left[g_{1}, \ldots, g_{n}\right]$
be an $n$-depth PGF kernel,
where $g_i,~i=1,\ldots,n$, is a Harris PGF with parameters $c_i$ and $r.$
The $n$-depth PGF kernel
$\mathcal{K}\left[g_{1}, \ldots, g_{n}\right]$
is the Harris kernel
$\mathcal{H}_{w_n,r}$
where
$w_{n}=\prod_{i=1}^{n}c_i .$
\end{proposition}

\begin{proof}
Equation~\ref{eq:harris_kernel_symm}
follows from equation~\ref{eq:harris_kernel} for even $r .$
\end{proof}

\begin{proof}
The proposition will be proved by induction.
For $n=1$,
according to equations~\ref{eq:kn:long:ve},~\ref{eq:harris_pgf} and~\ref{eq:harris_kernel},
notice that
\begin{equation*}
\mathcal{K}\left[g_1\right](x, z)=
g_{1}(\rho (x,z))=
\mathcal{H}_{w_1,r}(x,z).
\end{equation*}
Assuming that
\begin{equation}
\label{eq:harris_rec}
\mathcal{K}\left[g_{1}, \ldots, g_{n}\right] (x, z)=
\left(
w_n
\rho (x, z)^{-r}
- \left(
w_n
- 1\right)\right)^{-\frac{1}{r}} =
\mathcal{H}_{w_n,r}(x, z)
\end{equation}
holds for $n=j ,$
it will be shown that it also holds for $n=j+1$.
To this end,
\begin{alignat*}{2}
\mathcal{K}\left[g_{1},\ldots, g_{j+1}\right](x, z) &=
g_{j+1}\circ g_{j}\circ\dots\circ g_1
\left(\rho(x,z)\right) && \;
\mbox{See equation~\ref{eq:kn:long:ve}}\\
&=
g_{j+1}\circ\mathcal{K}\left[g_{1},\ldots, g_{j}\right](x, z)
&& \;
\mbox{See equation~\ref{eq:kn:long:ve}}
\\
&=
g_{j+1}\left(
\left(
w_j
\rho (x, z)^{-r}
- \left(
w_j
- 1\right)\right)^{-\frac{1}{r}}
\right)
&& \;
\mbox{See equation~\ref{eq:harris_rec}}
\\
&=
\left(
w_{j+1}
\rho (x, z)^{-r}
- \left(
w_{j+1}
- 1\right)\right)^{-\frac{1}{r}},
&& \;
\mbox{See equation~\ref{eq:harris_pgf}}
\end{alignat*}
which completes the proof.
\end{proof}


\begin{proposition}
\label{prop:lf_comp}
For any $n\in\mathbb{N},$ let
$\mathcal{K}\left[g_{1}, \ldots, g_{n}\right]$
be an $n$-depth PGF kernel,
where $g_i,~i=1,\ldots,n$, is a linear fractional PGF
with parameters $a_i$ and $b_i.$
The $n$-depth PGF kernel
$\mathcal{K}\left[g_{1}, \ldots, g_{n}\right]$
is the linear fractional kernel
$\mathcal{F}_{u_n,v_n}, $
where
$
u_n =  \amn{n}$ and
$v_n =
\sum_{i=1}^n u_{i-1} b_i .
$
\end{proposition}

\begin{proof}
The proposition will be proved by induction.
For $n=1$,
according to equations~\ref{eq:kn:long:ve},~\ref{eq:lf_pgf} and~\ref{eq:lf_kernel},
notice that
$\mathcal{K}\left[g_1\right](x, z)=
g_{1}(\rho (x,z))=
\mathcal{F}_{u_1,v_1}(x,z).
$
Assuming that
\begin{equation}
\label{eq:lf_rec}
\mathcal{K}\left[g_{1}, \ldots, g_{n}\right] (x, z)=
1 -
\left(
u_n \left(1-\rho (x, z)\right)^{-1}
+ v_n
\right)^{-1}
=
\mathcal{F}_{u_n,v_n}(x, z)
\end{equation}
holds for $n=j ,$
it will be shown that it also holds for $n=j+1$.
To this end,
\begin{alignat*}{2}
\mathcal{K}\left[g_{1},\ldots, g_{j+1}\right](x, z) &=
g_{j+1}\circ g_{j}\circ\dots\circ g_1
\left(\rho(x,z)\right) && \;
\mbox{See equation~\ref{eq:kn:long:ve}}\\
&=
g_{j+1}\circ\mathcal{K}\left[g_{1},\ldots, g_{j}\right](x, z)
&& \;
\mbox{See equation~\ref{eq:kn:long:ve}}
\\
&=
g_{j+1}\left(
1 -
\left(
u_j \left(1-\rho (x, z)\right)^{-1}
+ v_j
\right)^{-1}
\right)
&& \;
\mbox{See equation~\ref{eq:lf_rec}}
\\
&=
1 -
\left(
u_{j+1} \left(1-\rho (x, z)\right)^{-1}
+ v_{j+1}
\right)^{-1},
&& \;
\mbox{See equation~\ref{eq:lf_pgf}}
\end{alignat*}
which completes the proof.
\end{proof}


\begin{proposition}
\label{prop:theta_kernels}
A PGF kernel that arises from the $n$-fold iteration of a $\theta$ PGF
expresses in closed form
according to one of the nine cases of table~\ref{tab:thetaK}
depending on the values of parameters
$\theta,a,q,c,r$ of the $\theta$ PGF.
\end{proposition}

\begin{table}[t]
\centering
\begin{tabular}{c|l|l|l|l|l}
Case
&
\multicolumn{1}{c|}{$\mathcal{K}[(g)_n]$}
&
\multicolumn{1}{c|}{$\theta$}
&
\multicolumn{1}{c|}{$a$}
&
\multicolumn{1}{c|}{$q,\,c$}
&
\multicolumn{1}{c}{$r$} \\ \hline
1
&
$1-(a^n(1-\rho)^{-\theta }+(a^n-1)c)^{-1/\theta}$
&
$(0, 1]$
&
$(1,\infty)$
&
$(0,\infty)$
&
$1$ \\
2
&
$1-((1-\rho)^{-\theta }+nc)^{-1/\theta }$
&
$(0, 1]$
&
$1$
&
$(0,\infty)$
&
$1$ \\
3
&
$1-(a ^n(1-\rho)^{-\theta }+(1-a ^n)(1-q)^{-\theta })^{-1/\theta }$
&
$(0, 1]$
&
$(0, 1)$
&
$[0, 1)$
&
$1$ \\
4
&
$1-(1-q)^{1-a^n}(1-\rho)^{a^n}$
&
$0$
&
$(0, 1)$
&
$[0, 1)$
&
$1$ \\
5
&
$1-(a ^n(1-\rho)^{|\theta|}+(1-a ^n)(1-q)^{|\theta|})^{1/{|\theta|}}$
&
$(-1, 0)$
&
$(0, 1)$
&
$[0, 1)$
&
$1$ \\
6
&
$a^n\rho+(1-a^n)q$
&
$-1$
&
$(0, 1)$
&
$[0, 1]$
&
$1$ \\
7
&
$r-(a^n(r-\rho)^{-\theta}+(1-a^n)(r-q)^{-\theta})^{-1/\theta}$
&
$(0, 1]$
&
$(0, 1)$
&
$[0, 1]$
&
$(1, \infty)$ \\
8
&
$r-(r-q)^{1-a^n}(r-\rho)^{a^n}$
&
$0$
&
$(0, 1)$
&
$[0, 1]$
&
$(1, \infty)$ \\
9
&
$r-(a ^n(r-\rho)^{|\theta|}+(1-a ^n)(r-q)^{|\theta|})^{1/{|\theta|}}$
&
$(-1, 0)$
&
$(0, 1)$
&
$[0, 1]$
&
$(1, \infty)$ \\
\end{tabular}
\caption{PGF kernels
$\mathcal{K}\left[(g)_n\right]$,
where $g$ is a $\theta$ PGF and
$\rho (x, z)$ is the correlation between
$x\in\mathbb{S}^h$ and $z\in\mathbb{S}^h$.}
\label{tab:thetaK}
\end{table}

\begin{proof}
Closed-form expressions for $n$-fold iterations of a $\theta$ PGF $f$
are provided in~\citet[section 4]{sagitov2016}.
Thus, table~\ref{tab:thetaK} follows.
\end{proof}


\begin{proof}[Proof of proposition~\ref{prop:theta_kernel_ex}]
The proposition will be proved by induction,
starting from definition~\ref{def:comp_kernel_in_varying_env}.
For $n=1$,
notice that
$K\left[g_1\right](x,z)=
g_{1}(\rho (x,z))=\mathcal{T}_{\theta, c_1}(x,z)$.
Assuming that equation~\ref{eq:theta_kernel_ex} holds for $n=j$,
it will be shown that it also holds for $n=j+1$.
To this end,
\begin{alignat*}{2}
\mathcal{K}\left[g_{1},\ldots, g_{j+1}\right](x, z) &=
g_{j+1}\circ g_{j}\circ\dots\circ g_1
\left(\rho(x,z)\right) && \;
\mbox{See equation~\ref{eq:kn:long:ve}}\\
&=
f_{j+1}\circ\mathcal{K}_{1:j}(x, z)
&& \;
\mbox{See equation~\ref{eq:kn:long:ve}}
\\
&=
g_{j+1}\left(
1 - \left(
(1-\rho (x, z))^{-\theta}
+ \sum_{i = 1}^{j} c_i
\right)^{-\frac{1}{\theta}}
\right) && \;
\mbox{See equation~\ref{eq:theta_kernel_ex}}
\\
&=
1 - \left(
(1-\rho (x, z))^{-\theta}
+ \sum_{i = 1}^{j+1} c_i
\right)^{-\frac{1}{\theta}},
&& \;
\mbox{See equation~\ref{eq:PGFMain}}
\end{alignat*}
which completes the proof.
\end{proof}


\begin{proof}[Proof of proposition~\ref{prop:theta_kernel_ex_inf_depth}]
Taking the limit of equation~\ref{eq:theta_kernel_ex}
as $n$ tends to infinity
yields equation~\ref{eq:theta_k_limit}.
\end{proof}


\end{document}